%% file: main.tex
\documentclass[10pt,twocolumn,letterpaper]{article}

\usepackage[pagenumbers]{iccv} 

\usepackage{amsmath}
\usepackage{amssymb}
\usepackage{booktabs}
\usepackage{multirow}
\usepackage{multicol} 
\usepackage{graphicx}
\usepackage{pifont}
\usepackage{enumitem}
\usepackage{wrapfig}
\usepackage{colortbl}
\definecolor{usercolor}{RGB}{184,85,10}
\definecolor{assistantcolor}{RGB}{0,115,0}  
\let\oldding\ding
\renewcommand{\ding}[2][1]{\scalebox{#1}{\oldding{#2}}}
\usepackage[normalem]{ulem}
\usepackage{soul}
\usepackage{arydshln}
\usepackage{makecell}
\usepackage{tikz}

\definecolor{iccvblue}{rgb}{0.21,0.49,0.74}
\usepackage[pagebackref,breaklinks,colorlinks,allcolors=iccvblue]{hyperref}

\title{ExpVG: Investigating the Design Space of Visual Grounding \\
in Multimodal Large Language Model}

\author{Weitai Kang$^{1,}$\footnotemark[1], Weiming Zhuang$^{2}$, Zhizhong Li$^{2}$, Yan Yan$^{1}$, Lingjuan Lyu$^{2,}$\footnotemark[2] \\
$^1$University of Illinois Chicago, $^2$Sony AI
}

\usepackage[accsupp]{axessibility}
\begin{document}
\maketitle
\input{sec/0_abs}
\renewcommand{\thefootnote}{\fnsymbol{footnote}}
\footnotetext[1]{Work done during internship at Sony AI}
\footnotetext[2]{Corresponding author}
\input{sec/1_intro}

\input{sec/2_related}
\input{sec/4_exp}
\input{sec/5_con}
\clearpage
{\small
    \bibliographystyle{ieeenat_fullname}
    \bibliography{main}
}
\input{sec/X_suppl}


\end{document}

%% file: sec/0_abs.tex
\begin{abstract}
    Fine-grained multimodal capability in Multimodal Large Language Models (MLLMs) has emerged as a critical
    research direction, particularly for tackling the visual grounding (VG) problem. Despite the strong performance achieved by existing approaches, they often employ disparate design choices when fine-tuning MLLMs for VG, lacking systematic verification to support these designs.
    To bridge this gap,
    this paper presents a comprehensive study of various design choices that impact the VG performance of MLLMs.
    We conduct our analysis using LLaVA-1.5, which has been widely adopted in prior empirical studies of MLLMs. While more recent models exist, we follow this convention to ensure our findings remain broadly applicable and extendable to other architectures.
    We cover two key aspects: (1) exploring different visual grounding paradigms in MLLMs, identifying the most effective design, and providing our insights; and (2) conducting ablation studies on the design of grounding data to optimize MLLMs' fine-tuning for the VG task. 
    Finally, our findings
    contribute to a stronger MLLM for VG, achieving improvements of +5.6\% / +6.9\% / +7.0\% on RefCOCO/+/g over the
    LLaVA-1.5.
\end{abstract}

\begin{figure}
    \centering
    \includegraphics[width=0.4\textwidth]{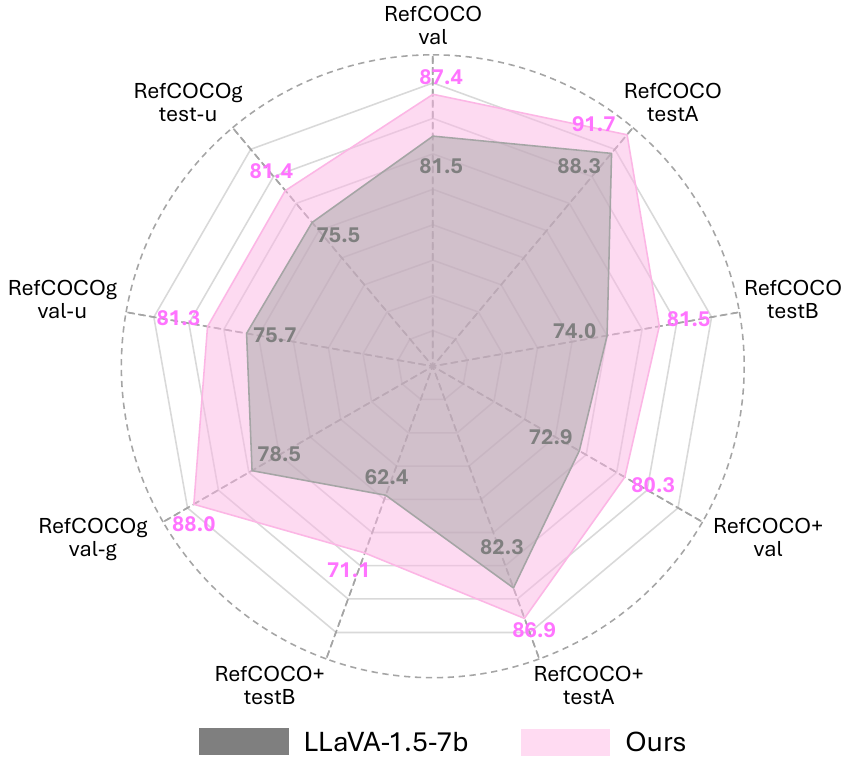}
    \vspace{-8pt}
    \caption{Using LLaVA-1.5-7B as the baseline, we conduct a comprehensive empirical study focusing on the design of visual grounding in Multimodal Large Language Models.
        We identify several improved designs while also ruling out certain potential alternatives. Finally, we integrate all the best designs, achieving a significant improvement over the baseline.}\label{radar_data}
    \vspace{-10pt}
\end{figure}

%% file: sec/1_intro.tex
\section{Introduction}
Visual grounding (VG) is a crucial vision-language learning task that aims to predict the location of an object in an image based on a given sentence~\cite{mao2016generation,plummer2015flickr30k,yu2016modeling, kazemzadeh2014referitgame,kang2025segvg,kang2024actress,kang2024visual}.
It facilitates
fine-grained cooperation between humans and AI agents in real-world scenarios~\cite{lei2024infant,kang2024intent3d,kang2025robin3dimproving3dlarge} and
benefits multimodal reasoning systems, such as visual question answering~\cite{gan2017vqs,wang2020general,shang2024llava} and image captioning~\cite{anderson2018bottom,chen2020say,you2016image}.
Early methods develop specialist models~\cite{deng2021transvg, kang2025segvg, yang2020improving,kang2024visual} with architectures tailored for visual grounding.
Extending beyond task-specific solutions, later approaches introduce unified models~\cite{li2022grounded, zhang2022glipv2, xiao2024florence, lu2022unified, lu2024unified, wang2022ofa, yang2022unitab, wang2024visionllm},
which benefit VG learning through integrating knowledge from multiple tasks.
Recently, researchers
explore incorporating VG capabilities into Multimodal Large Language Models (MLLMs)~\cite{you2023ferret, zhang2024ferret, chen2023shikra, liu2024improved, peng2023kosmos, xuan2024pink, he2024multi}.
Unlike unified models that require extensive fine-grained labels across tasks, MLLMs inherit strong reasoning abilities from LLMs~\cite{vicuna2023, touvron2023llama, touvron2023llama2} and general visual understanding from vision foundation models~\cite{radford2021learning, oquab2023dinov2, zhai2023sigmoid}, both of which
were trained on large-scale unlabeled data. This enables MLLMs to not only achieve strong VG performance after fine-tuning
but also support complex multimodal interactions,
such as multi-turn reasoning, making them promising for VG.

\begin{figure*}[t]
    \centering
    \includegraphics[width=0.9\textwidth]{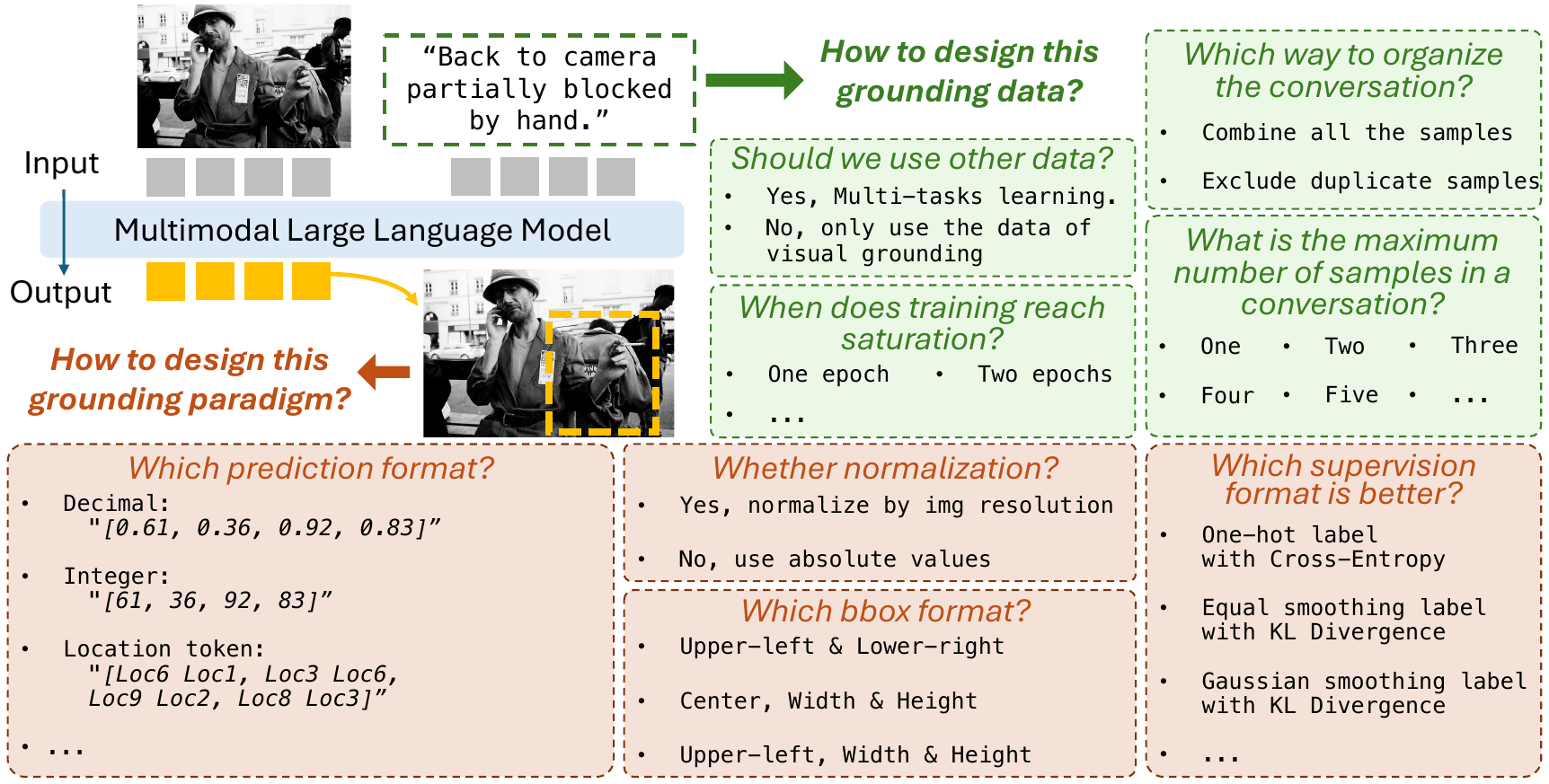}
    \vspace{-8pt}
    \caption{We conduct systematic ablation studies to explore different 
        design choices for building visual grounding ability in Multimodal Large Language Model. We 
        investigate two main 
        aspects: \colorbox[rgb]{0.984, 0.898, 0.839}{Grounding paradigm design} and \colorbox[rgb]{0.886, 0.941, 0.851}{Grounding data design}.}\label{fig2}
    \vspace{-10pt}
\end{figure*}


The idea of incorporating VG capability into MLLMs is inspired by Pix2Seq~\cite{chen2021pix2seq},
which pioneered the reformulation of bounding box regression as classification by discretizing values into different bins.
Despite the rapid progress in adapting MLLMs for VG, current studies often present disparate design choices, lacking a comprehensive experimental justification.
Meanwhile, existing empirical studies on the design of MLLMs~\cite{karamcheti2024prismatic, shi2024eagle, laurenccon2024matters} primarily focus on training recipes~\cite{karamcheti2024prismatic, laurenccon2024matters}, model structure choices~\cite{shi2024eagle, laurenccon2024matters}, and
textual benchmarks~\cite{karamcheti2024prismatic, laurenccon2024matters} such as
visual question answering and image captioning.
However, these efforts are insufficient to fully characterize MLLMs' potential in fine-grained multimodal reasoning,
as they have not thoroughly explored the highly diverse VG design in MLLMs, which is a crucial part of MLLMs to serve as generalist models.
Moreover, within the VG community, advancing MLLMs
to excel in this domain~\cite{you2023ferret, chen2023shikra, peng2023kosmos, xuan2024pink, he2024multi} remains a topic requiring further exploration.

To address these gaps, we systematically investigate different design choices for building VG ability in MLLMs,
which complements previous studies.
Following the convention of prior empirical studies on MLLMs~\cite{karamcheti2024prismatic, shi2024eagle}, we ground our research by
using LLaVA-1.5~\cite{liu2024improved},
one of the most
popular works in MLLM development,
as the baseline.
As shown in \cref{fig2}, we systematically study on the design choices in the grounding paradigm design and the grounding data design for VG in MLLMs.

Specifically,
we highlight four key contributions.
First, for grounding paradigm design,
\textbf{we systematically examine how bounding boxes (bbox) should be represented}, \eg, whether normalization is necessary.
Notably, our exploration extends beyond existing MLLM designs (\eg, different bbox prediction formats) to incorporate new alternative approaches inspired from non-MLLM domains (\eg, different bbox supervision formats).
Second, for grounding data design,
\textbf{we identify the effective data configurations}, and reveal that using pure VG data and deduplicated conversational samples leads to improved learning efficiency.
Third,
\textbf{we provide insights into those better design choices}.
We introduce a similarity-based correlation metric to quantify the enhancements on the spatial semantics of coordinate tokens brought by training with the one-hot label and cross-entropy loss.
This metric
could be a useful tool for analyzing MLLM's VG behavior in future research.
Finally, by incorporating the optimal design choices identified in our findings, \textbf{we achieve substantial improvements over the baseline, LLaVA-1.5}, as shown in~\cref{radar_data}.
Crucially, we not only reveal the effective designs in MLLMs—encompassing both existing and unexplored approaches—but also uncover the ineffective designs, despite their prior adoption and perceived potential.
Our findings offer clear guidelines for the future development of MLLM-based VG\@.

%% file: sec/2_related.tex
\section{Background}

\subsection{Related Work}
\paragraph{Classification Paradigm in Fine-Grained Task.}\label{related_work}
Many works adopt the classification paradigm for fine-grained
visual recognition, including those in object detection, visual grounding, and human pose estimation.
Pix2seq~\cite{chen2021pix2seq}, OFA~\cite{wang2022ofa}, and KOSMOS-2~\cite{peng2023kosmos} discretize image locations and introduce extra vocabularies into language modeling to represent bounding box coordinates. Methods such as Shikra~\cite{chen2023shikra}, LLaVA-1.5~\cite{liu2024improved}, and Pink~\cite{xuan2024pink} directly treat the
textual representation of bounding boxes as prediction targets and classify each digit in those decimal values
using language modeling.
Each training sample is organized as a conversation, transforming the visual grounding problem into a question-answering format.
Ferret~\cite{you2023ferret}, Ferret-v2~\cite{zhang2024ferret}, and MM1.5~\cite{zhang2024mm1} convert decimals into integers by quantizing each coordinate into one of 1000 discrete bins and classifying
the textual representation of these integer values
using language modeling.
The values are not normalized by image resolution.
In
human pose estimation, SimCC~\cite{li2022simcc} discretizes each keypoint location into discrete bins and classifies the bins.
It introduces label smoothing to account for the spatial relevance of adjacent bins.
In addition to explicitly generating bounding box coordinates, some MLLMs employ an implicit representation for referring image segmentation~\cite{lai2024lisa,ren2024pixellm,rasheed2024glamm}.
For example, Lisa~\cite{lai2024lisa} uses a special language token to indicate an object and decodes the segmentation mask from the hidden state of this token.
Yet, due to differences in model structure, task, data, and training recipe, the optimal design
for visual grounding 
in MLLMs remains an open question.

\paragraph{Empirical Study of MLLMs.}
As MLLMs
continue to advance the field of vision-language learning, researchers have begun to explore their design space by empirical studies.
Prismatic~\cite{karamcheti2024prismatic} investigates the training recipe of MLLMs based on LLaVA-1.5~\cite{liu2024improved}.
Eagle~\cite{shi2024eagle} explores the design space for MLLMs, focusing on vision encoders and input resolutions via question-answering benchmarks.
Idefics2~\cite{laurenccon2024matters} conducts extensive experiments on pretrained models, architectures, data, and training methods, benchmarking them on question-answering and captioning tasks.
However, none of these works provide empirical studies on the design choices for visual grounding in MLLMs.

\subsection{Preliminaries}\label{preliminary}

\paragraph{Model Architecture.}
For an input image, we use a pretrained CLIP-ViT-L-336px~\cite{radford2021learning} as the visual encoder.
Its output visual features are projected into the LLM's word embedding space via a two-layer MLP (vision-language connector), producing a sequence of visual tokens.
The corresponding texts for questions and answers are tokenized and projected into
text tokens.
The visual tokens and text tokens are then concatenated into a single sequence.
Given the initial part of the sequence, the LLM, Vicuna-7B-v1.5~\cite{vicuna2023}, predicts the next token based on the preceding tokens. Unless specified otherwise, we use the cross-entropy loss.
Only the tokens corresponding to the answer text are considered as the supervised learning targets.

\paragraph{Implementation.}\label{implementation}
We use the
official codebase from LLaVA-1.5~\cite{llava-code} and retain its training hyperparameters to ensure reproducibility.
We adopt the pretrained vision-language connector from LLaVA-1.5 and fine-tune both the connector and the LLM\@.
We extract visual grounding data from LLaVA-1.5's 665K multimodal instruction-tuning examples to conduct systematic ablation studies.
In total, we extract 112,370 visual grounding conversational samples originating from RefCOCO/+/g~\cite{kazemzadeh2014referitgame, mao2016generation} and Visual Genome~\cite{krishna2017visual}, with additional annotations provided by the LLaVA-1.5 authors.
Each sample consists of an image and multiple rounds of question-answering for the visual grounding task.
The remaining samples, including visual question-answering data from \citet{liu2024visual,hudson2019gqa,sidorov2020textcaps} and language-only question-answering data from \citet{sharegpt}, are used in studying the synergistic effect of multitask learning
in \cref{synergy}.
Without specified otherwise, we use only visual grounding data for training by default.

\paragraph{Evaluation Suite.}
We follow common practices~\cite{deng2021transvg, kang2024actress, yang2022improving, kang2025segvg, yu2018mattnet, kang2024visual, yang2020improving, kang2024intent3d} to evaluate visual grounding performance on RefCOCO/+/g~\cite{kazemzadeh2014referitgame, mao2016generation}.
RefCOCO emphasizes brief descriptions with spatial references, RefCOCO+ focuses solely on appearance-based descriptions, and RefCOCOg centers on extended, detailed descriptions.
We evaluate bounding box prediction accuracy, considering a predicted bounding box correct if its Intersection over Union (IoU) with the ground-truth bounding box exceeds 0.5.

We emphasize that this is an empirical study aimed at laying the foundation for the future development of visual grounding in MLLMs.
\emph{To ensure the fairness of our ablation studies and the rigor of our findings, we intentionally refrain from comparing with other MLLMs, as disparities in data utilization, training scope, and model parameters would provide limited meaningful insights in our systematic experiments.}

%% file: sec/4_exp.tex
\section{Grounding Paradigm Design}\label{Exp}

In this section, we investigate the grounding paradigm designs of MLLMs,
including the prediction format in \cref{predicition_format}, the normalization type in \cref{normalization}, the supervision format in \cref{supervision_format}, and the bounding box format in \cref{bbox_format}.

\subsection{Prediction Format}\label{predicition_format}

Following the classification paradigms outlined in \cref{related_work}, we examine five candidate formats for bounding box targets, covering both explicit representations in various formats~\cite{liu2024improved,you2023ferret,chen2021pix2seq,peng2023kosmos} and implicit approaches~\cite{lai2024lisa,ren2024pixellm,he2024multi}.

\begin{figure}[t]
    \centering
    \includegraphics[width=0.4\textwidth]{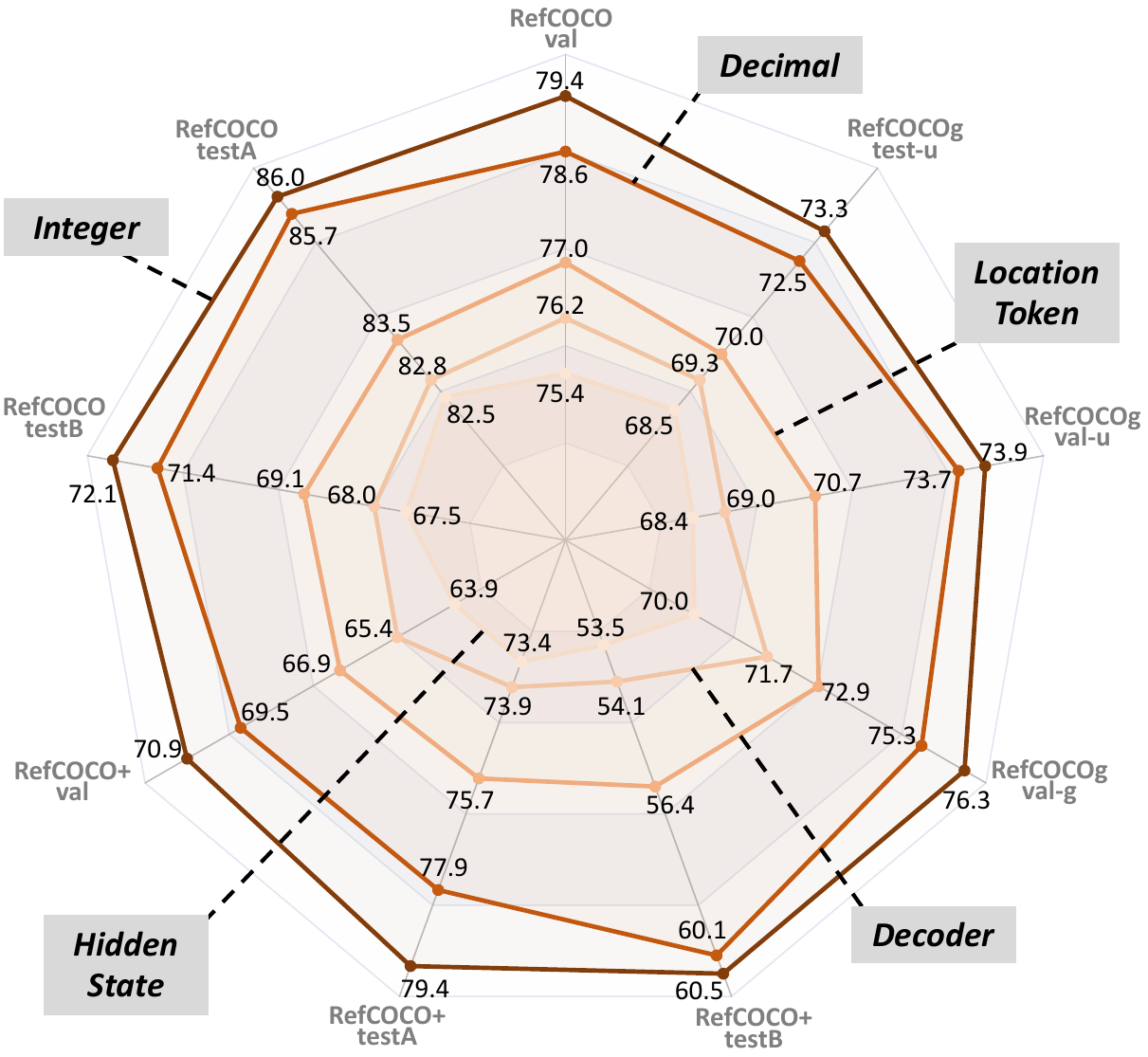}
    \vspace{-9pt}
    \caption{We explore the choices on the prediction format for MLLM's visual grounding paradigm design.
        We find that explicit prediction formats, like \textit{decimal, integer}, and \textit{location token} formats universally outperform implicit prediction formats (\textit{hidden state, decoder}).
        And \textit{integer} format provides the best performance.}\label{fig3}
\end{figure}

\noindent 1) \textit{Decimal format}.
Following \citet{chen2023shikra, liu2024improved, xuan2024pink}, the bbox values range from~0 to~1 after normalization
by the image resolution.
For instance, the MLLM may be trained to predict the string ``\textit{[0.17, 0.23, 0.8, 0.65]}''.

\noindent 2) \textit{Integer format}.
Following approaches in~\cite{you2023ferret, zhang2024ferret, zhang2024mm1}, we convert the decimals to integers by multiplying them by 100, yielding integer strings such as ``\textit{[17, 23, 80, 65]}''.
The conversion is performed on normalized ranges, which ensures consistency with other formats in our comparison.
However, note that the aforementioned methods adopt the absolute values without normalization by the image resolution.
We will discuss the normalization design in \cref{normalization}.

\noindent 3) \textit{Location token format}.
To explore the use of new vocabularies to represent location, as in \citet{chen2021pix2seq, peng2023kosmos, wang2022ofa}, we discretize the image coordinates into $n$ bins.
We then add extra vocabulary tokens to represent these bins.
In alignment with the original vocabulary, which only has ten words from 0 to 9 to represent digits, we add ten new location words, from ``\textit{Loc0}'' to ``\textit{Loc9}''.
For instance, the MLLM predicts two tokens, ``\textit{Loc1} \textit{Loc7}'',
to indicate the 17th bin.
We set $n$ to 101 to match the numerical precision of the previous format.

\noindent 4) \textit{Hidden state format}.
Inspired by Lisa~\cite{lai2024lisa}, PixelLM~\cite{ren2024pixellm}, and AnyRef~\cite{he2024multi}, we introduce a special token, ``\texttt{<Det>}'',  to the vocabulary. The model needs to predict this token in language modeling and decode its hidden state to predict the bounding box using traditional object detection losses. 
To evaluate the effectiveness of this format in a simplified setting,
we use a three-layer MLP to decode the hidden state.

\noindent 5) \textit{Decoder format}.
To further investigate the ideas from Lisa, PixelLM, and AnyRef while preserving approximately equivalent model parameters for a fair comparison, we augment the \textit{hidden state} format
by adding three extra transformer layers as a decoder.
This allows the hidden state to perform cross-attention on the vision features extracted by the vision encoder before predicting the bounding box.

\noindent \textbf{Results.}
Under the experimental setup outlined in \cref{implementation}, where models are trained for one epoch on visual grounding data, we observe two key findings, as illustrated in \cref{fig3}:
1) Within a comparable model parameter scope, explicit prediction formats—namely \textit{decimal, integer}, and \textit{location token} formats—are more effective for training than implicit formats, such as \textit{hidden state} and \textit{decoder} formats.
2) Among the explicit formats, the \textit{integer} format achieves the best performance, significantly surpassing the \textit{location token} format and slightly outperforming the \textit{decimal} format.

\begin{figure}[t]
    \centering
    \includegraphics[width=0.45\textwidth]{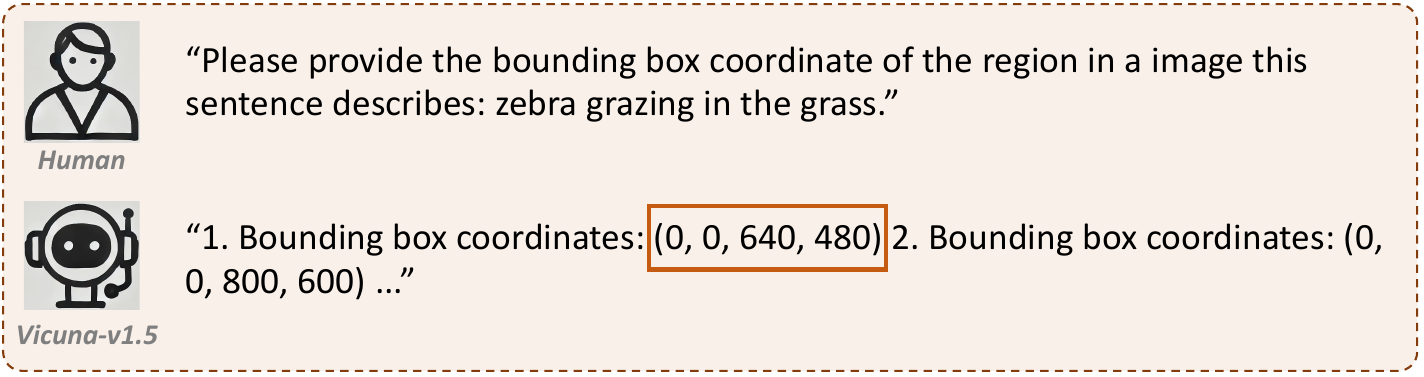}
    \vspace{-9pt}
    \caption{To investigate why the \textit{integer} format is superior in MLLM's visual grounding paradigm, we examine the behavior of the pretrained LLM model, Vicuna-v1.5~\cite{vicuna2023},
        from which our MLLM is fine-tuned. We find that the pretrained LLM inherently utilizes the \textit{integer} format, making it more natural and effective to continue using this format during fine-tuning.
    }\label{fig4}
\end{figure}

\noindent \textbf{Discussion.}
To understand why the \textit{integer} format outperforms others in MLLM's visual grounding paradigm, we examine the behavior of the pretrained LLM model, Vicuna-v1.5, from which our MLLM is fine-tuned.
As shown in \cref{fig4}, when we prompt Vicuna-v1.5 with a visual grounding task using only the description (without an image), we find that the pretrained LLM inherently adopts the \textit{integer} format.
This suggests that using the \textit{integer} format during fine-tuning may better align with the model's pretrained knowledge, making the training more effective.

\subsection{Normalization Type}\label{normalization}
Given the varying choices in previous works \citep{chen2023shikra, liu2024improved, you2023ferret, zhang2024ferret, zhang2024mm1} on whether the bounding box should be normalized by image resolution, we conduct a comprehensive comparison by evaluating each normalization type—\textit{normalized} or \textit{unnormalized}—across three prediction formats: \textit{decimal}, \textit{integer}, and \textit{decoder}.

\noindent 1) \textit{Normalized type}.
Following \citet{chen2023shikra} and \citet{liu2024improved}, the bounding box is normalized by the image resolution.
This setting is adopted in many existing visual grounding methods~\cite{deng2021transvg, kang2025segvg, kang2024visual, kang2024actress}.
For example, given a 640×640 image and a bounding box of ``\textit{[32, 32, 320, 320]}'', the \textit{normalized} type in \textit{decimal} format is ``\textit{[0.05, 0.05, 0.5, 0.5]}'', and in \textit{integer} format, it is ``\textit{[5, 5, 50, 50]}''.

\noindent 2) \textit{Unnormalized type}.
In contrast, \citet{you2023ferret, zhang2024ferret, zhang2024mm1} adopt an \textit{unnormalized} type, where the bounding box retains its absolute values without normalization.
In this case, for the same example, its \textit{decimal} format is obtained by dividing by a fixed maximum value (\eg, 1280), resulting in ``\textit{[0.025, 0.025, 0.25, 0.25]}'', while its \textit{integer} format remains as the absolute values: ``\textit{[32, 32, 320, 320]}''.

\begin{figure}[t]
    \centering
    \includegraphics[width=0.48\textwidth]{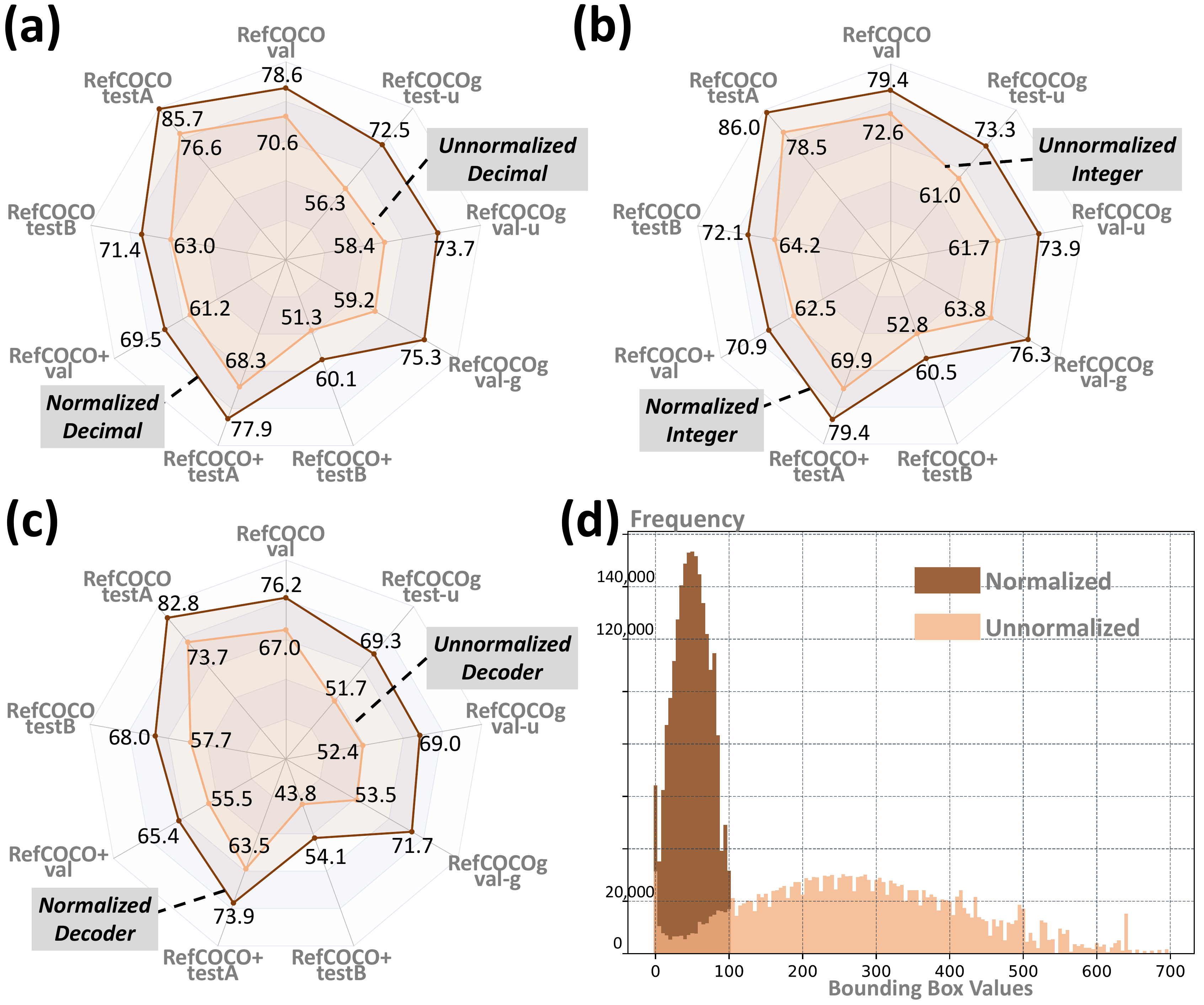}
    \vspace{-16pt}
    \caption{
        (a,b,c): Performance comparison of normalization types -- \textit{normalized} and \textit{unnormalized} types under three prediction formats -- \textit{decimal}, \textit{integer}, and \textit{decoder} formats.
        (d): Frequency comparison of \textit{normalized} and \textit{unnormalized} types.
        The bar chart shows the frequency of unique location values, illustrating the concentrated distribution of \textit{normalized} types and the long-tailed distribution of \textit{unnormalized} types.}\label{fig5}
    \vspace{-6pt}
\end{figure}

\noindent \textbf{Results.}
As shown in \cref{fig5}(a,b,c), under a fair experimental setting and comprehensive evaluation, we find that the more common choice, \textit{normalized type}, consistently outperforms the \textit{unnormalized} type.

\noindent \textbf{Discussion.}
Using the \textit{integer} format, we analyze the frequency distribution of location values in the training dataset.
As shown in the histogram in \cref{fig5}(d), the distribution for the \textit{normalized} type is notably more concentrated, as normalization enables a coordinate token to represent various absolute location values across different image resolutions.
In contrast, the \textit{unnormalized} type exhibits a broader range of location values, leading to increased variability and less efficient training compared to the \textit{normalized} type.

To quantitatively evaluate the long-tailed distribution, we compute \texttt{\small Excess Kurtosis} (\texttt{\small EK}), which measures the sharpness and tail heaviness of a distribution.
A higher \texttt{\small EK} (positive value) indicates heavier tails and a stronger long-tail effect, while a lower \texttt{\small EK} (negative value) suggests lighter tails and fewer extreme values.
For the \textit{normalized} type, the \texttt{\small EK} is \texttt{\small -0.7595}, indicating relatively light tails.
In contrast, the \textit{unnormalized} type has a \texttt{\small EK} of \texttt{\small 0.6611}, signifying a more pronounced long-tail effect.
We further validate these observations using the \textit{decimal} format, which yields consistent results: for the \textit{normalized} type, \texttt{\small EK} is \texttt{\small -0.7595}; for the \textit{unnormalized} type, \texttt{\small EK} is \texttt{\small 0.6677}.
In summary, the \textit{normalized} type produces a more balanced distribution with lighter tails, which is advantageous for mitigating long-tail effects in training.

\begin{table}[t]
    \centering
    \setlength{\tabcolsep}{0.6mm}
    \resizebox{0.48\textwidth}{!}{
        \begin{tabular}{ccccccccccc}
            \Xhline{1.pt}  
            \addlinespace[0.1cm]
            \multirow{2}{*}{Format} & \multicolumn{3}{c}{RefCOCO}      & \multicolumn{3}{c}{RefCOCO+}     & \multicolumn{3}{c}{RefCOCOg}     & Ave.$\downarrow$                                                                                                                                                                                                                                           \\
                                    & \textit{val}                     & \textit{testA}                   & \textit{testB}                   & \textit{val}                     & \textit{testA}                   & \textit{testB}                   & \textit{val-g}                   & \textit{val-u}                   & \textit{test-u}                  & Rank                                     \\
            \hline
            \multicolumn{11}{c}{\textit{Using Decimal Prediction Format}}                                                                                                                                                                                                                                                                                                                                 \\
            $One$-$hot$             & \color[rgb]{0.6, 0.2, 0.1}{78.6} & \color[rgb]{0.6, 0.2, 0.1}{85.7} & \color[rgb]{0.6, 0.2, 0.1}{71.4} & \color[rgb]{0.6, 0.2, 0.1}{69.5} & \color[rgb]{0.6, 0.2, 0.1}{77.9} & \color[rgb]{0.6, 0.2, 0.1}{60.1} & \color[rgb]{0.6, 0.2, 0.1}{75.3} & \color[rgb]{0.6, 0.2, 0.1}{73.7} & \color[rgb]{0.6, 0.2, 0.1}{72.5} & \colorbox[rgb]{0.984, 0.898, 0.839}{1.0} \\
            $Equal$                 & 77.9                             & 85.0                             & 70.6                             & 68.5                             & 77.8                             & 58.1                             & 74.2                             & 72.0                             & 72.2                             & 3.2                                      \\
            $Gaussian$              & 78.4                             & 85.3                             & 70.9                             & 69.0                             & \color[rgb]{0.6, 0.2, 0.1}{77.9} & 59.3                             & 74.8                             & 72.4                             & 72.2                             & 2.0                                      \\
            $Gaussian^{W}$          & 77.8                             & \color[rgb]{0.6, 0.2, 0.1}{85.7} & 70.0                             & 68.3                             & 77.1                             & 57.4                             & 74.7                             & 72.1                             & 71.6                             & 3.4                                      \\
            \hline
            \multicolumn{11}{c}{\textit{Using Integer Prediction Format}}                                                                                                                                                                                                                                                                                                                                 \\
            $One$-$hot$             & 79.4                             & 86.0                             & \color[rgb]{0.6, 0.2, 0.1}{72.1} & \color[rgb]{0.6, 0.2, 0.1}{70.9} & \color[rgb]{0.6, 0.2, 0.1}{79.4} & \color[rgb]{0.6, 0.2, 0.1}{60.5} & \color[rgb]{0.6, 0.2, 0.1}{76.3} & 73.9                             & \color[rgb]{0.6, 0.2, 0.1}{73.3} & \colorbox[rgb]{0.984, 0.898, 0.839}{1.4} \\
            $Equal$                 & 79.4                             & 86.2                             & 71.0                             & 70.0                             & 78.8                             & 59.5                             & 76.2                             & \color[rgb]{0.6, 0.2, 0.1}{74.0} & 73.2                             & 2.2                                      \\
            $Gaussian$              & \color[rgb]{0.6, 0.2, 0.1}{79.8} & \color[rgb]{0.6, 0.2, 0.1}{86.4} & 71.0                             & 70.1                             & 79.2                             & 59.9                             & 76.0                             & 73.6                             & 72.9                             & 2.1                                      \\
            $Gaussian^{W}$          & 73.5                             & 77.3                             & 64.2                             & 64.4                             & 71.5                             & 54.3                             & 60.0                             & 67.8                             & 66.1                             & 4.0                                      \\
            \Xhline{1.pt}  
        \end{tabular}}
    \vspace{-8pt}
    \caption{Performance of different supervision formats across various benchmarks.
        $Gaussian^{W}$ indicates \textit{weighted Gaussian label smoothing}.
        The \textit{one-hot} format consistently achieves the highest rank, followed by the \textit{Gaussian label smoothing} format, in both \textit{decimal} and \textit{integer} prediction formats.
        The average rank is computed by averaging rankings across different benchmarks.
    }\label{tab1}
\end{table}

\subsection{Supervision Format}\label{supervision_format}
MLLMs use cross-entropy loss with one-hot encoded label as supervision.
In human pose estimation, SimCC~\cite{li2022simcc}, a method that also adopts the classification paradigm, incorporates \textit{label smoothing} to address annotation noise.
Given its demonstrated effectiveness in the pixel-level task, we hypothesize that \textit{label smoothing} might similarly enhance the performance of MLLM in visual grounding.
Therefore, we investigate this potential for a thorough and comprehensive empirical study.

\noindent 1) \textit{One-hot}.
MLLMs employ an autoregressive language modeling approach, wherein the ground-truth of each token is represented by the one-hot encoding and the objective is to minimize the cross-entropy loss.

\noindent 2) \textit{Equal label smoothing}.
SimCC~\cite{li2022simcc} adopts the equal label smoothing technique in keypoint location prediction.
Specifically, the ground-truth label is represented as a probability distribution, where the correct category is assigned a high probability value (\texttt{\small 0.9}), while the remaining categories receive a uniform low probability (\texttt{\small 0.1}).
This technique aims to mitigate noise introduced by human annotations.
The corresponding loss function is the KL-Divergence loss.

\noindent 3) \textit{Gaussian label smoothing}.
Instead of uniformly assigning low probabilities to all incorrect categories, SimCC~\cite{li2022simcc} further introduces \textit{Gaussian label smoothing}, where the ground-truth probability distribution follows a Gaussian distribution centered at the correct category.
This approach ensures that locations closer to the ground truth receive higher probabilities, effectively capturing the varying loss intensity due to spatial proximity.
We set the standard deviation to \texttt{\small 0.5}, as it yields the best performance.

\noindent 4) \textit{Weighted Gaussian label smoothing}.
We further consider a crucial factor specific to visual grounding in MLLMs.
When predicting a value autoregressively, errors in different digits of the value have varying impacts.
For example, given a target value of ``0.17,'' an error of one unit in the tenths place results in an overall error of ``0.1,'' whereas the same error in the hundredths place leads to only ``0.01.''
To account for this, we assign lower variance to higher-magnitude digits.
Specifically, for the \textit{decimal} format, we set the standard deviations for the ones, tenths, and hundredths places to \texttt{\small 0.1}, \texttt{\small 0.5}, and \texttt{\small 0.7}, respectively.
For the \textit{integer} format, we set the standard deviations for the tens and ones places to \texttt{\small 0.1} and \texttt{\small 0.5}, respectively.

\noindent \textbf{Results.}
As shown in \cref{tab1}, the most commonly used \textit{one-hot} supervision format outperforms others across most benchmarks, regardless of whether the prediction format is \textit{decimal} or \textit{integer}.
To quantify the relative effectiveness of different supervision formats, we rank them across multiple benchmarks and compute their average rank.
The \textit{one-hot} format achieves the highest rank, followed by the \textit{Gaussian label smoothing} format. 

\begin{figure}[t]
    \centering
    \includegraphics[width=0.48\textwidth]{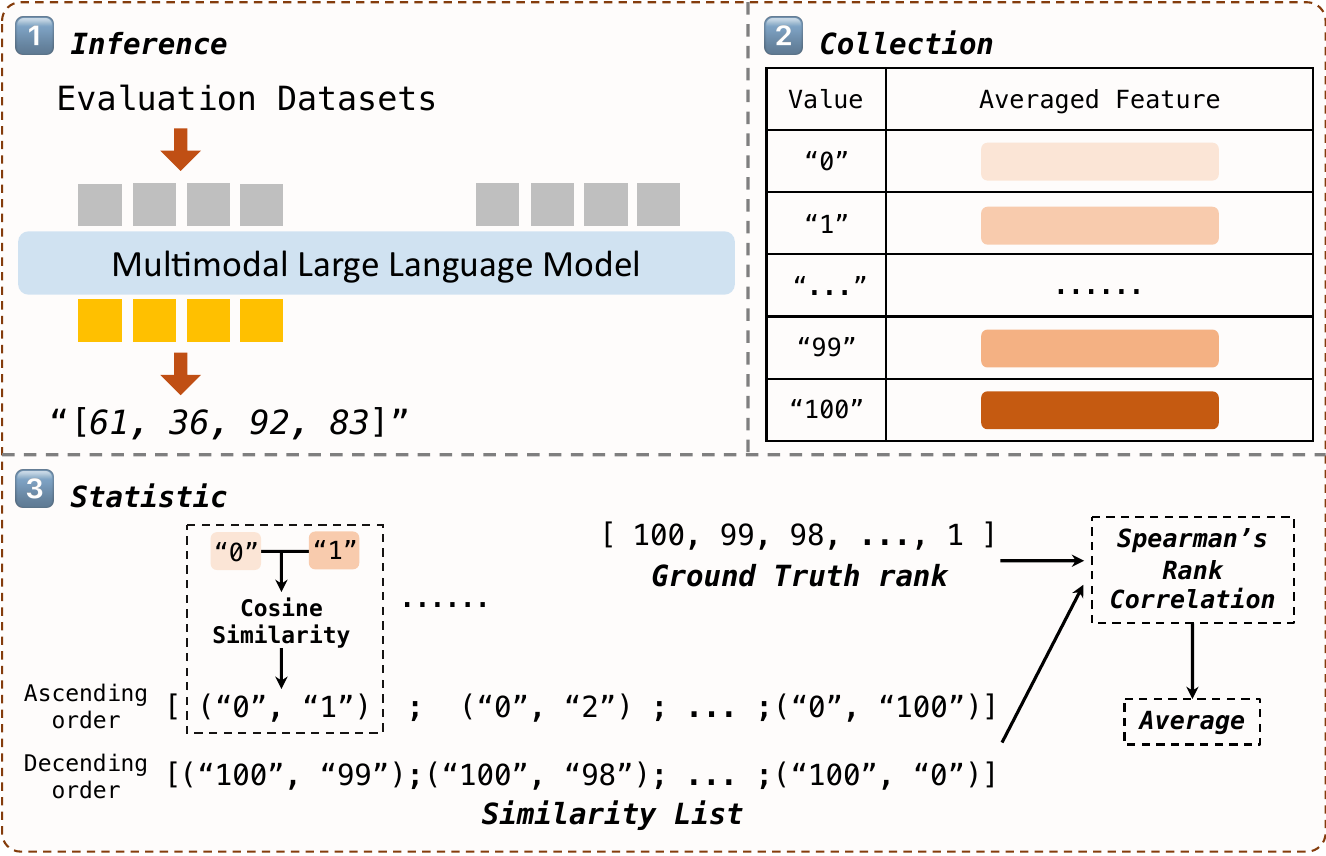}
    \vspace{-10pt}
    \caption{Illustration of the similarity-based correlation metric.
        After collecting the feature of different coordinate values,
        we compute token similarities by selecting 0 and 100 as anchor points and measuring cosine similarity between coordinates and anchors in ascending and descending order, respectively.
        We then use Spearman’s Rank Correlation (rho) to evaluate the alignment between the two similarity lists and the ground-truth rank list.
        Finally, we average the two rho results from both lists.
    }\label{fig6}
    \vspace{-8pt}
\end{figure}

\noindent \textbf{Discussion.}
To investigate whether coordinate tokens trained with the \textit{one-hot} format encode spatial semantics, we propose a similarity-based correlation metric.
Our hypothesis is that coordinates closer in value should exhibit more similar token representations.
As illustrated in \cref{fig6}, we follow a three-step process:
1) We run inference on all evaluation datasets.
2) For each coordinate in the model’s output, we extract its corresponding token features from the last layer and compute their average.
3) We select 0 and 100 as anchor points and compute the cosine similarity between each coordinate and the anchor—one in ascending order and the other in descending order—thus generating two similarity lists.
If our hypothesis holds perfectly, these lists should follow a strictly decreasing order.
To quantitatively evaluate this hypothesis, we construct a ground-truth ranking list from 100 to 1 in descending order, corresponding to the expected behavior under a perfectly valid hypothesis.
We compute Spearman’s Rank Correlation between the similarity lists and the ground-truth ranking, respectively, and then average the results to obtain our final similarity-based correlation metric.
A value closer to 1 supports the hypothesis, indicating that ``closer coordinates exhibit higher token similarity.'' Conversely, a value approaching -1 provides strong evidence against the hypothesis.
Using the \textit{integer} format, we obtain a correlation score of \texttt{\small 0.6396} for the \textit{one-hot} supervision format and \texttt{\small 0.4680} for the \textit{Gaussian label smoothing} format.
This suggests that training with the \textit{one-hot} format effectively encodes spatial semantics, with a stronger effect compared to the \textit{Gaussian label smoothing} format.

\begin{table}[t]
    \centering
    \setlength{\tabcolsep}{0.6mm}
    \resizebox{0.48\textwidth}{!}{
        \begin{tabular}{ccccccccccc}
            \Xhline{1.pt}  
            \addlinespace[0.1cm]
            \multirow{2}{*}{Format}      & \multicolumn{3}{c}{RefCOCO}      & \multicolumn{3}{c}{RefCOCO+}     & \multicolumn{3}{c}{RefCOCOg}     & Ave.$\downarrow$                                                                                                                                                                                                                                           \\
                                         & \textit{val}                     & \textit{testA}                   & \textit{testB}                   & \textit{val}                     & \textit{testA}                   & \textit{testB}                   & \textit{val-g}                   & \textit{val-u}                   & \textit{test-u}                  & Rank                                     \\
            \hline
            \multicolumn{11}{c}{\textit{Using Decimal Prediction Format}}                                                                                                                                                                                                                                                                                                                                      \\
            $X_{1}, Y_{1}, X_{2}, Y_{2}$ & \color[rgb]{0.6, 0.2, 0.1}{78.6} & \color[rgb]{0.6, 0.2, 0.1}{85.7} & \color[rgb]{0.6, 0.2, 0.1}{71.4} & 69.5                             & 77.9                             & \color[rgb]{0.6, 0.2, 0.1}{60.1} & 75.3                             & \color[rgb]{0.6, 0.2, 0.1}{73.7} & 72.5                             & \colorbox[rgb]{0.984, 0.898, 0.839}{1.4} \\
            $X_{c}, Y_{c}, W, H$         & 78.5                             & 85.4                             & 70.6                             & 68.8                             & \color[rgb]{0.6, 0.2, 0.1}{78.2} & 58.7                             & \color[rgb]{0.6, 0.2, 0.1}{75.4} & 72.8                             & 71.8                             & 2.4                                      \\
            $X_{1}, Y_{1}, W, H$         & \color[rgb]{0.6, 0.2, 0.1}{78.6} & 85.5                             & 71.2                             & \color[rgb]{0.6, 0.2, 0.1}{69.8} & 77.9                             & 59.1                             & 75.0                             & 72.8                             & \color[rgb]{0.6, 0.2, 0.1}{72.9} & 1.8                                      \\
            \hline
            \multicolumn{11}{c}{\textit{Using Integer Prediction Format}}                                                                                                                                                                                                                                                                                                                                      \\
            $X_{1}, Y_{1}, X_{2}, Y_{2}$ & 79.4                             & \color[rgb]{0.6, 0.2, 0.1}{86.0} & 72.1                             & \color[rgb]{0.6, 0.2, 0.1}{70.9} & \color[rgb]{0.6, 0.2, 0.1}{79.4} & \color[rgb]{0.6, 0.2, 0.1}{60.5} & \color[rgb]{0.6, 0.2, 0.1}{76.3} & \color[rgb]{0.6, 0.2, 0.1}{73.9} & 73.3                             & \colorbox[rgb]{0.984, 0.898, 0.839}{1.4} \\
            $X_{c}, Y_{c}, W, H$         & 79.6                             & 85.8                             & 70.3                             & 69.9                             & 77.8                             & 59.4                             & 75.6                             & 73.1                             & 73.3                             & 2.4                                      \\
            $X_{1}, Y_{1}, W, H$         & \color[rgb]{0.6, 0.2, 0.1}{79.7} & 85.7                             & \color[rgb]{0.6, 0.2, 0.1}{72.2} & 70.3                             & 79.0                             & 60.3                             & 75.4                             & 72.2                             & \color[rgb]{0.6, 0.2, 0.1}{73.5} & 2.0                                      \\
            \Xhline{1.pt}  
        \end{tabular}}
    \vspace{-8pt}
    \caption{Performance comparison of different bounding box formats across benchmarks.
        The $X_1, Y_1, X_2, Y_2$ format consistently achieves the highest average rank in both \textit{decimal} and \textit{integer} formats, indicating its superiority over formats from traditional visual grounding methods and datasets in MLLMs. 
    }\label{tab2}
\end{table}
\subsection{Bounding Box Format}\label{bbox_format}
Now we discuss the choices on the format of bounding box based on \textit{decimal} and \textit{integer} prediction format.

\noindent 1) {\(X_1, Y_1, X_2, Y_2\)}.
It is a widely-used bounding box format in autoregressive approach~\cite{chen2021pix2seq,liu2024improved,you2023ferret}, where (\(X_1, Y_1\)) is the upper-left and (\(X_2, Y_2\)) is the lower-right coordinates.

\noindent 2) \(X_c, Y_c, W, H\).
Some visual grounding methods, \eg \citet{kang2025segvg,deng2021transvg}, use center coordinates (\(X_c, Y_c\)) together with width and height (\(W, H\)) to indicate a bounding box.
This format has not been adopted in previous MLLMs.
Thus, we explore this design in MLLMs.

\noindent 3) \(X_1, Y_1, W, H\).
It is the default bounding box format for benchmark datasets~\citep{kazemzadeh2014referitgame, mao2016generation}, where (\(X_1, Y_1\)) is upper-left coordinates and (\(W, H\)) is width and height.
This alternative has not been used in previous MLLMs and we explore this design in MLLMs.

\noindent \textbf{Results.}
As shown in \cref{tab2}, the $X_{1}, Y_{1}, X_{2}, Y_{2}$ bounding box format consistently achieves the highest average rank in both \textit{decimal} and \textit{integer} formats.
This demonstrates that the bounding box format used in existing visual grounding methods and the format used in datasets are less effective than the format used in autoregressive approaches.

\section{Grounding Data Design}\label{trainrecipe}
We explore the design choices for grounding data, including the synergistic effects of multitask training in \cref{synergy}, different ways to organize conversational data in \cref{organization}, and the scaling property on training time in \cref{epoch}.
Unless specified otherwise, we adopt the combination of the best performing choices as our grounding paradigm for the subsequent studies, specifically, the \textit{normalized integer} in \(X_1, Y_1, X_2, Y_2\) format with \textit{one-hot} supervision.

\subsection{Synergistic Effect}\label{synergy}
Multitask training is widely recognized as an effective approach to enhancing performance through its synergistic effects~\cite{liu2024improved, zhang2024ferret}.
However, it remains an open question whether multitask training is effective in attaining visual grounding ability within the same training budget.
Here, we compare three settings.

\noindent 1) \textit{Visual grounding data}.
The baseline is trained on our default data which consists of pure visual grounding data, as described in the \cref{preliminary}.

\noindent 2) \textit{Visual grounding + VQA data}.
To investigate the synergistic effect from multitask training, we incorporate the remaining VQA data from LLaVA-1.5's 665K multimodal instruct tuning examples.

\noindent 3) \textit{Scaled visual grounding data}.
To provide a fair
comparison by using the same training budget, we scale the \textit{visual grounding data} by randomly sampling and duplicating instances from the dataset until the total number of samples matches the \textit{visual grounding + VQA data}.

\noindent \textbf{Results.}
As shown in \cref{tab3}, \textit{scaled visual grounding data} and \textit{visual grounding + VQA data} demonstrate that, under the same training cost, using only visual grounding data is more effective than incorporating various VQA tasks in multitask training.
Notably, \textit{scaled visual grounding data} is constructed by simply duplicating samples from \textit{visual grounding data} without introducing new unique samples.
Therefore, given a fixed training budget, we believe that increasing the diversity in visual ground data is more effective than the synergistic effect from multitask training for building MLLM's visual grounding ability.

\begin{table}[t]
    \centering
    \setlength{\tabcolsep}{0.6mm}
    \resizebox{0.48\textwidth}{!}{
        \begin{tabular}{cccccccccc}
            \Xhline{1.pt}  
            \addlinespace[0.1cm]
            \multirow{2}{*}{Format} & \multicolumn{3}{c}{RefCOCO}                       & \multicolumn{3}{c}{RefCOCO+}                      & \multicolumn{3}{c}{RefCOCOg}                                                                                                                                                                                                                                                                                                                                              \\
                                    & \textit{val}                                      & \textit{testA}                                    & \textit{testB}                                    & \textit{val}                                      & \textit{testA}                                    & \textit{testB}                                    & \textit{val-g}                                    & \textit{val-u}                                    & \textit{test-u}                                   \\
            \hline
            Visual                  & \multirow{2}{*}{79.4}                             & \multirow{2}{*}{86.0}                             & \multirow{2}{*}{72.1}                             & \multirow{2}{*}{70.9}                             & \multirow{2}{*}{79.4}                             & \multirow{2}{*}{60.5}                             & \multirow{2}{*}{76.3}                             & \multirow{2}{*}{73.9}                             & \multirow{2}{*}{73.3}                             \\
            grounding data          &                                                   &                                                   &                                                   &                                                   &                                                   &                                                   &                                                   &                                                   &                                                   \\
            \hdashline
            Visual grounding        & \multirow{2}{*}{81.6}                             & \multirow{2}{*}{87.8}                             & \multirow{2}{*}{74.8}                             & \multirow{2}{*}{73.6}                             & \multirow{2}{*}{81.2}                             & \multirow{2}{*}{62.5}                             & \multirow{2}{*}{78.4}                             & \multirow{2}{*}{76.1}                             & \multirow{2}{*}{76.3}                             \\
            + VQA data              &                                                   &                                                   &                                                   &                                                   &                                                   &                                                   &                                                   &                                                   &                                                   \\
            \hdashline
            Scaled visual           & \multirow{2}{*}{\color[rgb]{0.0, 0.5, 0.0}{85.2}} & \multirow{2}{*}{\color[rgb]{0.0, 0.5, 0.0}{90.3}} & \multirow{2}{*}{\color[rgb]{0.0, 0.5, 0.0}{79.3}} & \multirow{2}{*}{\color[rgb]{0.0, 0.5, 0.0}{76.8}} & \multirow{2}{*}{\color[rgb]{0.0, 0.5, 0.0}{84.3}} & \multirow{2}{*}{\color[rgb]{0.0, 0.5, 0.0}{67.8}} & \multirow{2}{*}{\color[rgb]{0.0, 0.5, 0.0}{85.4}} & \multirow{2}{*}{\color[rgb]{0.0, 0.5, 0.0}{78.2}} & \multirow{2}{*}{\color[rgb]{0.0, 0.5, 0.0}{78.7}} \\
            grounding data          &                                                   &                                                   &                                                   &                                                   &                                                   &                                                   &                                                   &                                                   &                                                   \\
            \Xhline{1.pt}  
        \end{tabular}}
    \caption{Comparison of different training data configurations to assess the synergistic effect.
        The results show that under the same training cost, using only visual grounding data (\textit{scaled visual grounding data}) outperforms multitask training with both visual grounding and VQA data (\textit{visual grounding + VQA data}).
    }\label{tab3}
\end{table}

\subsection{Conversation Organization}\label{organization}
In MLLMs~\cite{liu2024improved, liu2024visual}, each training data is structured as a multi-round conversation.
Specifically, given an image, multiple image-related question-answering pairs are sequentially concatenated as one data sample.
This structure influences the in-context learning in two key aspects, \textit{duplicated annotations} and the \textit{maximum number of conversation rounds}.

\noindent \textbf{Duplicated Annotations}.
Given a fixed bounding box for an object, visual grounding data may include multiple referential sentences describing the object.
Consequently, in conversational data, the answer (bounding box) for the current round's question may have already appeared in a previous round, resulting in ground truth leakage and thereby weakens the training sample.
To investigate this consideration, we perform the following ablation study:

\noindent 1) \textit{Original conversational data}.
We use the original conversational data of visual grounding as described in \cref{preliminary} as the baseline.

\noindent 2) \textit{Deduplicated conversational data}.
We extract question-answering pairs from the original data where answers are duplicated and create new conversational data from these pairs.
This process is repeated iteratively until no datum contain duplicated answers, resulting in 161,827 samples.

\noindent 3) \textit{Scaled original data}.
Since the \textit{deduplicated conversational data} has 49,457 more data than the \textit{original}, the training steps has been increased.
To ensure a fair ablation study, we scale the \textit{original} data by randomly repeating samples to match the number of samples in the \textit{deduplicated} data.

\noindent \textbf{Results.}
\cref{tab4} lists the result of the ablation study on both \textit{decimal} and \textit{integer} prediction formats.
The \textit{deduplicated conversational data} consistently achieves superior performance, underscoring the importance of eliminating the duplicated answers, which prevents ground truth leakage and improves the quality of training data.

\begin{table}[t]
    \centering
    \setlength{\tabcolsep}{0.6mm}
    \resizebox{0.48\textwidth}{!}{
        \begin{tabular}{cccccccccc}
            \Xhline{1.pt}  
            \addlinespace[0.1cm]
            \multirow{2}{*}{Format} & \multicolumn{3}{c}{RefCOCO}      & \multicolumn{3}{c}{RefCOCO+}     & \multicolumn{3}{c}{RefCOCOg}                                                                                                                                                                                                                       \\
                                    & \textit{val}                     & \textit{testA}                   & \textit{testB}                   & \textit{val}                     & \textit{testA}                   & \textit{testB}                   & \textit{val-g}                   & \textit{val-u}                   & \textit{test-u}                  \\
            \hline
            \multicolumn{10}{c}{\textit{Using Decimal Prediction Format}}                                                                                                                                                                                                                                                                                                                                      \\
            Original                & 78.6                             & 85.7                             & 71.4                             & 69.5                             & 77.9                             & 60.1                             & 75.3                             & 73.7                             & 72.5                             \\
            Deduplicate             & \color[rgb]{0.0, 0.5, 0.0}{83.3} & \color[rgb]{0.0, 0.5, 0.0}{88.8} & \color[rgb]{0.0, 0.5, 0.0}{76.9} & \color[rgb]{0.0, 0.5, 0.0}{75.1} & \color[rgb]{0.0, 0.5, 0.0}{82.7} & \color[rgb]{0.0, 0.5, 0.0}{66.8} & \color[rgb]{0.0, 0.5, 0.0}{80.1} & \color[rgb]{0.0, 0.5, 0.0}{77.3} & \color[rgb]{0.0, 0.5, 0.0}{77.0} \\
            Scaled Original         & 82.0                             & 88.5                             & 76.2                             & 73.3                             & 81.7                             & 64.2                             & 78.3                             & 75.8                             & 75.4                             \\
            \hline
            \multicolumn{10}{c}{\textit{Using Integer Prediction Format}}                                                                                                                                                                                                                                                                                                                                      \\
            Original                & 79.4                             & 86.0                             & 72.1                             & 70.9                             & 79.4                             & 60.5                             & 76.3                             & 73.9                             & 73.3                             \\
            Deduplicate             & \color[rgb]{0.0, 0.5, 0.0}{83.9} & \color[rgb]{0.0, 0.5, 0.0}{89.2} & \color[rgb]{0.0, 0.5, 0.0}{77.4} & \color[rgb]{0.0, 0.5, 0.0}{75.1} & \color[rgb]{0.0, 0.5, 0.0}{83.3} & \color[rgb]{0.0, 0.5, 0.0}{66.6} & \color[rgb]{0.0, 0.5, 0.0}{80.6} & \color[rgb]{0.0, 0.5, 0.0}{77.6} & \color[rgb]{0.0, 0.5, 0.0}{77.5} \\
            Scaled Original         & 82.7                             & 89.0                             & 76.3                             & 74.2                             & 82.1                             & 64.7                             & 79.0                             & 76.3                             & 76.1                             \\
            \Xhline{1.pt}  
        \end{tabular}}
        \vspace{-6pt}
    \caption{Ablation study on the impact of duplicated annotations in conversational data.
        Results show that \textit{deduplicated conversational data} consistently outperforms others across both \textit{decimal} and \textit{integer} prediction formats, emphasizing the importance of removing duplicated answer samples to enhance data quality.
    }\label{tab4}
\end{table}

\begin{table}[!h]
    \centering
    \setlength{\tabcolsep}{0.6mm}
    \resizebox{0.48\textwidth}{!}{
        \begin{tabular}{ccccccccccc}
            \Xhline{1.pt}
            \addlinespace[0.1cm]
            Maximum              & \multicolumn{3}{c}{RefCOCO}                       & \multicolumn{3}{c}{RefCOCO+}                      & \multicolumn{3}{c}{RefCOCOg}                      & Ave.$\downarrow$                                                                                                                                                                                                                                                                                                                                                              \\
            Round                & \textit{val}                                      & \textit{testA}                                    & \textit{testB}                                    & \textit{val}                                      & \textit{testA}                                    & \textit{testB}                                    & \textit{val-g}                                    & \textit{val-u}                                    & \textit{test-u}                                   & Rank                                                  \\
            \multirow{2}{*}{One} & \multirow{2}{*}{85.8}                             & \multirow{2}{*}{90.8}                             & \multirow{2}{*}{80.0}                             & \multirow{2}{*}{77.8}                             & \multirow{2}{*}{84.9}                             & \multirow{2}{*}{68.9}                             & \multirow{2}{*}{81.7}                             & \multirow{2}{*}{78.7}                             & \multirow{2}{*}{79.1}                             & \multirow{2}{*}{8.7}                                  \\
                                 &                                                   &                                                   &                                                   &                                                   &                                                   &                                                   &                                                   &                                                   &                                                   &                                                       \\
            \hdashline
            Two                  & \multirow{2}{*}{86.5}                             & \multirow{2}{*}{90.6}                             & \multirow{2}{*}{80.3}                             & \multirow{2}{*}{78.5}                             & \multirow{2}{*}{\color[rgb]{0.0, 0.5, 0.0}{86.0}} & \multirow{2}{*}{70.2}                             & \multirow{2}{*}{84.7}                             & \multirow{2}{*}{80.3}                             & \multirow{2}{*}{80.6}                             & \multirow{2}{*}{4.6}                                  \\
            (Scaled)             &                                                   &                                                   &                                                   &                                                   &                                                   &                                                   &                                                   &                                                   &                                                   &                                                       \\
            \hdashline
            Three                & \multirow{2}{*}{\color[rgb]{0.0, 0.5, 0.0}{86.7}} & \multirow{2}{*}{91.0}                             & \multirow{2}{*}{\color[rgb]{0.0, 0.5, 0.0}{80.8}} & \multirow{2}{*}{79.5}                             & \multirow{2}{*}{85.9}                             & \multirow{2}{*}{\color[rgb]{0.0, 0.5, 0.0}{71.0}} & \multirow{2}{*}{85.7}                             & \multirow{2}{*}{79.7}                             & \multirow{2}{*}{\color[rgb]{0.0, 0.5, 0.0}{80.7}} & \multirow{2}{*}{\colorbox[rgb]{0.786, 1, 0.751}{2.9}} \\
            (Scaled)             &                                                   &                                                   &                                                   &                                                   &                                                   &                                                   &                                                   &                                                   &                                                   &                                                       \\
            \hdashline
            Four                 & \multirow{2}{*}{86.6}                             & \multirow{2}{*}{91.0}                             & \multirow{2}{*}{80.0}                             & \multirow{2}{*}{79.2}                             & \multirow{2}{*}{85.5}                             & \multirow{2}{*}{69.3}                             & \multirow{2}{*}{86.4}                             & \multirow{2}{*}{\color[rgb]{0.0, 0.5, 0.0}{80.5}} & \multirow{2}{*}{\color[rgb]{0.0, 0.5, 0.0}{80.7}} & \multirow{2}{*}{3.9}                                  \\
            (Scaled)             &                                                   &                                                   &                                                   &                                                   &                                                   &                                                   &                                                   &                                                   &                                                   &                                                       \\
            \hdashline
            Five                 & \multirow{2}{*}{86.6}                             & \multirow{2}{*}{90.9}                             & \multirow{2}{*}{79.5}                             & \multirow{2}{*}{\color[rgb]{0.0, 0.5, 0.0}{79.6}} & \multirow{2}{*}{\color[rgb]{0.0, 0.5, 0.0}{86.0}} & \multirow{2}{*}{69.7}                             & \multirow{2}{*}{86.6}                             & \multirow{2}{*}{80.4}                             & \multirow{2}{*}{80.4}                             & \multirow{2}{*}{3.8}                                  \\
            (Scaled)             &                                                   &                                                   &                                                   &                                                   &                                                   &                                                   &                                                   &                                                   &                                                   &                                                       \\
            \hdashline
            Six                  & \multirow{2}{*}{86.4}                             & \multirow{2}{*}{\color[rgb]{0.0, 0.5, 0.0}{91.1}} & \multirow{2}{*}{79.3}                             & \multirow{2}{*}{79.2}                             & \multirow{2}{*}{85.6}                             & \multirow{2}{*}{70.0}                             & \multirow{2}{*}{87.6}                             & \multirow{2}{*}{79.5}                             & \multirow{2}{*}{80.5}                             & \multirow{2}{*}{4.4}                                  \\
            (Scaled)             &                                                   &                                                   &                                                   &                                                   &                                                   &                                                   &                                                   &                                                   &                                                   &                                                       \\
            \hdashline
            Seven                & \multirow{2}{*}{86.0}                             & \multirow{2}{*}{90.5}                             & \multirow{2}{*}{79.7}                             & \multirow{2}{*}{79.0}                             & \multirow{2}{*}{85.5}                             & \multirow{2}{*}{69.8}                             & \multirow{2}{*}{87.5}                             & \multirow{2}{*}{80.0}                             & \multirow{2}{*}{80.2}                             & \multirow{2}{*}{6.2}                                  \\
            (Scaled)             &                                                   &                                                   &                                                   &                                                   &                                                   &                                                   &                                                   &                                                   &                                                   &                                                       \\
            \hdashline
            Eight                & \multirow{2}{*}{85.6}                             & \multirow{2}{*}{\color[rgb]{0.0, 0.5, 0.0}{91.1}} & \multirow{2}{*}{79.0}                             & \multirow{2}{*}{78.4}                             & \multirow{2}{*}{85.5}                             & \multirow{2}{*}{69.3}                             & \multirow{2}{*}{87.6}                             & \multirow{2}{*}{79.7}                             & \multirow{2}{*}{79.6}                             & \multirow{2}{*}{6.7}                                  \\
            (Scaled)             &                                                   &                                                   &                                                   &                                                   &                                                   &                                                   &                                                   &                                                   &                                                   &                                                       \\
            \hdashline
            Nine                 & \multirow{2}{*}{86.1}                             & \multirow{2}{*}{91.0}                             & \multirow{2}{*}{79.3}                             & \multirow{2}{*}{79.3}                             & \multirow{2}{*}{85.6}                             & \multirow{2}{*}{69.3}                             & \multirow{2}{*}{\color[rgb]{0.0, 0.5, 0.0}{87.8}} & \multirow{2}{*}{79.2}                             & \multirow{2}{*}{80.0}                             & \multirow{2}{*}{5.3}                                  \\
            (Scaled)             &                                                   &                                                   &                                                   &                                                   &                                                   &                                                   &                                                   &                                                   &                                                   &                                                       \\
            \hdashline
            Ten                  & \multirow{2}{*}{86.4}                             & \multirow{2}{*}{90.7}                             & \multirow{2}{*}{79.2}                             & \multirow{2}{*}{79.3}                             & \multirow{2}{*}{85.7}                             & \multirow{2}{*}{68.8}                             & \multirow{2}{*}{87.5}                             & \multirow{2}{*}{80.1}                             & \multirow{2}{*}{79.8}                             & \multirow{2}{*}{6.1}                                  \\
            (Scaled)             &                                                   &                                                   &                                                   &                                                   &                                                   &                                                   &                                                   &                                                   &                                                   &                                                       \\
            \Xhline{1.pt}
        \end{tabular}}
    \vspace{-8pt}
    \caption{Ablation study on the maximum number of conversation rounds.
        Setting the maximum number to three achieves the best balance, enhancing the model's reasoning capability across different aspects of the image while preventing excessive ground truth leakage that could overly simplify training.
    }\label{tab5}
    \vspace{-4pt}
\end{table}

\begin{table}[t]
    \centering
    \setlength{\tabcolsep}{0.6mm}
    \resizebox{0.48\textwidth}{!}{
        \begin{tabular}{ccccccccccc}
            \Xhline{1.pt}
            \addlinespace[0.1cm]
            \multirow{2}{*}{Epoch} & \multicolumn{3}{c}{RefCOCO}      & \multicolumn{3}{c}{RefCOCO+}     & \multicolumn{3}{c}{RefCOCOg}     & Ave.$\downarrow$                                                                                                                                                                                                                                       \\
                                   & \textit{val}                     & \textit{testA}                   & \textit{testB}                   & \textit{val}                     & \textit{testA}                   & \textit{testB}                   & \textit{val-g}                   & \textit{val-u}                   & \textit{test-u}                  & Rank                                 \\
            \hline
            One                    & 84.7                             & 89.8                             & 78.5                             & 76.2                             & 84.0                             & 67.9                             & 81.1                             & 78.1                             & 78.4                             & 7.0                                  \\
            Two                    & \color[rgb]{0.0, 0.5, 0.0}{87.4} & \color[rgb]{0.0, 0.5, 0.0}{91.7} & \color[rgb]{0.0, 0.5, 0.0}{81.7} & 80.1                             & 86.1                             & \color[rgb]{0.0, 0.5, 0.0}{71.3} & 85.4                             & 80.5                             & 80.7                             & 2.4                                  \\
            Three                  & \color[rgb]{0.0, 0.5, 0.0}{87.4} & 91.6                             & 81.1                             & 80.1                             & 86.2                             & \color[rgb]{0.0, 0.5, 0.0}{71.3} & 87.5                             & \color[rgb]{0.0, 0.5, 0.0}{81.3} & \color[rgb]{0.0, 0.5, 0.0}{81.4} & 2.1                                  \\
            Four                   & \color[rgb]{0.0, 0.5, 0.0}{87.4} & \color[rgb]{0.0, 0.5, 0.0}{91.7} & 81.5                             & \color[rgb]{0.0, 0.5, 0.0}{80.3} & \color[rgb]{0.0, 0.5, 0.0}{86.9} & 71.1                             & 88.0                             & \color[rgb]{0.0, 0.5, 0.0}{81.3} & \color[rgb]{0.0, 0.5, 0.0}{81.4} & \colorbox[rgb]{0.786, 1, 0.751}{1.8} \\
            Five                   & 87.0                             & 91.1                             & 80.2                             & 80.0                             & 86.1                             & 71.2                             & 89.0                             & 81.1                             & 80.3                             & 4.0                                  \\
            Six                    & 86.7                             & 90.6                             & 80.5                             & 79.7                             & 86.1                             & 70.6                             & \color[rgb]{0.0, 0.5, 0.0}{89.4} & 80.4                             & 80.4                             & 4.2                                  \\
            Seven                  & 86.2                             & 90.8                             & 80.3                             & 79.1                             & 85.7                             & 69.6                             & 89.2                             & 79.6                             & 80.4                             & 5.1                                  \\
            \Xhline{1.pt}
        \end{tabular}}
    \vspace{-8pt}
    \caption{Ablation study on the number of training epochs.
        The model achieves peak performance at four epochs, after which additional training yields negative returns.
    }\label{tab6}
    \vspace{-4pt}
\end{table}

\begin{figure}[t]
    \centering
    \includegraphics[width=0.46\textwidth]{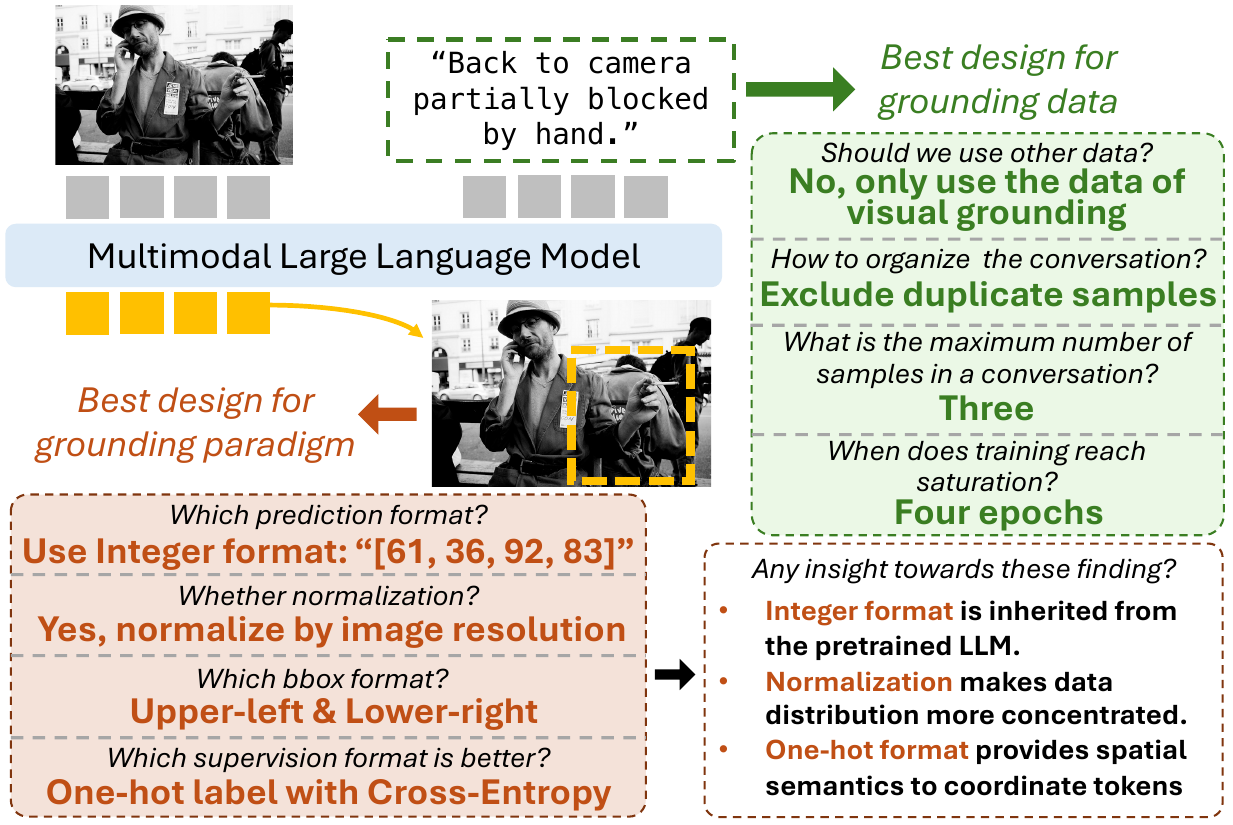}
    \vspace{-8pt}
    \caption{An overview of the best designs for training visual grounding in MLLM\@.
        We highlight key contributions, including optimal design choices on grounding paradigm and grounding data.
        We emphasize several insights towards our findings.
    }\label{fig8}
    \vspace{-8pt}
\end{figure}

\noindent \textbf{Maximum Number of Conversation Rounds}.
Increasing the number of conversation rounds introduces greater challenges in comprehensively reasoning about different aspects of the image.
However, it also inevitably reveals more identified objects in the answers (\ie, ground truth bounding boxes), thereby reducing the difficulty of grounding in subsequent rounds of question-answering.
Therefore, we conduct a comprehensive ablation study to determine the optimal maximum number of conversation rounds, ranging from one to ten.
We use \textit{deduplicated} data and split the conversational data into subsets by truncating conversations at the specified maximum round limit.
For example, if the maximum round is set to 5, a 10-round conversation is split into two samples, each containing 5 rounds.
To ensure a fair comparison, we scale each dataset by randomly repeating samples to the same amount of the one round data setting.

\noindent \textbf{Results.}
As shown in \cref{tab5}, setting the \textit{maximum number of conversations} to 3 achieves the best balance.
This setting enhances the model's ability to handle reasoning challenges across different aspects of the image while preventing excessive leakage of ground truth information, which could otherwise reduce training difficulty.

\subsection{Scaling Training Time}\label{epoch}
To study the scaling property on training time when training visual grounding in MLLMs, we vary the training epochs from one to seven.
Based on previous section, we use \textit{deduplicated} data with \textit{maximum three conversation rounds}.

\noindent \textbf{Results.}
As shown in \cref{tab6}, training for 2 to 4 epochs results in strong performance, with 4 epochs yielding the best results.
Performance improves significantly up to the second epoch, with marginal gains thereafter.
Beyond the fourth epoch, additional training offers diminishing returns.

%% file: sec/5_con.tex
\vspace{-10pt}
\section{Summary \& Conclusion}\label{finalversion}
\vspace{-6pt}
We present an empirical study on visual grounding in MLLMs based on LLaVA-1.5.
Our study identifies optimal design choices and discusses the gained insights.
As shown in \cref{fig8}, we propose to adopt the \textit{normalized integer} format with the \textit{upper-left and lower-right} bounding box representation, and train with \textit{one-hot} labels using the cross-entropy loss.
The \textit{integer} format aligns with the pretrained LLM, while \textit{normalization} mitigates long-tailed distributions.
Our similarity-based correlation metric reveals that \textit{one-hot} supervision enhances spatial semantics.
For grounding data design, training on visual grounding data outperforms multitask training and removing duplicated answers improves learning.
The optimal conversation round is three, and optimal training epoch is four.
Integrating these designs, we significantly surpass LLaVA-1.5 by +5.6\% / +6.9\% / +7.0\% on RefCOCO/+/g.
We believe our study lays a foundation for future research in advancing MLLM's VG capability.

%% file: sec/X_suppl.tex
\clearpage
\setcounter{page}{1}
\maketitlesupplementary

\section{Impact of data volume}
We analyze the impact by using 50\%, 100\%, and 150\% of the data in Tab.~\ref{tab7}. On the aspect of normalization type, the normalized type outperforms unnormalized type regardless of data volume -- the conclusion is consistent with our finding in the main paper.

\begin{table}[t]
\centering
\resizebox{\linewidth}{!}{
\begin{tabular}{lccccc}
\toprule
data & normalize & refcoco val & refcoco+ val & refcocog val-g \\
\midrule
50\% & Unnormalized & 55.8 & 48.3 & 49.0 \\
50\% & Normalized & 65.4 & 55.0 & 56.6 \\
\hdashline
100\% & Unnormalized & 72.6 & 62.5 & 63.8 \\
100\% & Normalized & 79.4 & 70.9 & 76.3 \\
\hdashline
150\% & Unnormalized & 75.0 & 65.0 & 66.0 \\
150\% & Normalized & 81.1 & 69.9 & 71.6 \\
\bottomrule
\end{tabular}
}
\caption{The impact of data volume in our empirical studies.
    }\label{tab7}
\end{table}

\section{Generalization to other MLLM architecture and baseline selection}
We provide experiments using Qwen2.5-VL architecture by training 1 epoch on our VG data in Tab.~\ref{tab8} to demonstrate the generalization of our findings to stronger MLLMs. The results show that on the prediction format and normalization type aspects, the \textit{integer + normalizaed} type is better, which is consistent with our findings in the main paper. 

\begin{table}[t]
\centering
\resizebox{\linewidth}{!}{
\begin{tabular}{lccccc}
\toprule
 Prediction & Normalization & refcoco val & refcoco+ val & refcocog val-g \\
\midrule
 Decimal & Unnormalized & 60.2	& 49.7	& 48.9 \\
 Integer & Unnormalized & 61.1	& 51.0	& 50.5 \\
 Integer & Normalized & 65.2	& 54.9	& 55.2\\
\bottomrule
\end{tabular}
}
\caption{Generalization to other MLLM architecture about our empirical studies.
    }\label{tab8}
\end{table}

\section{Clarification on Cross-Factor}
We have provided several important cross-factor studies in the main paper:
(1) \textit{Prediction format × supervision strategy} We contain 8 combinations with \{decimal, integer\} formats × \{one-hot, equal, gaussian, gaussian\_w\} strategies. 
(2) {\textit{Prediction format × normalization type}} We contain 6 combinations with \{decimal, integer, decoder\} formats × \{normalized, unnormalized\} types.  
(3) {\textit{Prediction format × bbox format}} We contain 6 combinations with \{decimal, integer\} formats × \{[X1,Y1,X2,Y2], [Xc,Yc,W,H], [X1,Y1,W,H]\} formats. 
(4) {\textit{Prediction format × conversation organization}} We contain 6 combinations with \{decimal, integer\} formats × \{original, deduplicate, scaled original\} organizations.  
In all cases, the best combinations consistently align with our findings: {integer} + {one-hot} + {normalized} + {[X1,Y1,X2,Y2]} + {deduplicate}.